\documentclass{article}

\usepackage{PRIMEarxiv}

\usepackage[utf8]{inputenc}
\usepackage[T1]{fontenc}
\usepackage{hyperref}
\usepackage{url}
\usepackage{booktabs}
\usepackage{amsfonts}
\usepackage{amssymb}
\usepackage{nicefrac}
\usepackage{microtype}
\usepackage{graphicx}
\usepackage{multirow}
\usepackage{array}
\usepackage{siunitx}
\usepackage{xcolor}
\usepackage{listings}
\usepackage{algorithm}
\usepackage{algpseudocode}

\definecolor{codegreen}{rgb}{0,0.6,0}
\definecolor{codegray}{rgb}{0.5,0.5,0.5}
\definecolor{codepurple}{rgb}{0.58,0,0.82}
\definecolor{backcolour}{rgb}{0.97,0.97,0.97}

\title{\includegraphics[width=0.99\textwidth]{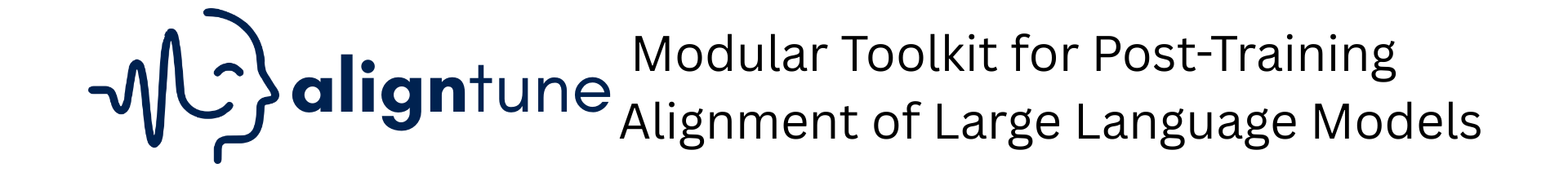}}

\author{
    R E Zera Marveen Lyngkhoi\thanks{Co-First Authorship} ,  Chirag Chawla\footnotemark[1]   \thanks{Department of Chemistry, Indian Institute of Technology (Banaras Hindu University) Varanasi, India.}   \thanks{work done during internship at Lexsi Labs} , Pratinav Seth\footnotemark[1] , \\ 
  Utsav Avaiya, Soham Bhattacharjee, Mykola Khandoga, Rui Yuan, \\
  Vinay Kumar Sankarapu \\
  \affiliation{Lexsi Labs}\\
}

\setabstract{%
Post-training alignment is central to deploying large language models (LLMs), yet practical workflows remain split across backend-specific tools and ad-hoc glue code, making experiments hard to reproduce. We identify backend interference, reward fragmentation, and irreproducible pipelines as key obstacles in alignment research. We introduce \emph{AlignTune}, a modular toolkit exposing a unified interface for supervised fine-tuning (SFT) and RLHF-style optimization with interchangeable TRL and Unsloth backends. AlignTune standardizes configuration, provides an extensible reward layer (rule-based and learned), and integrates evaluation over standard benchmarks and custom tasks. By isolating backend-specific logic behind a single factory boundary, AlignTune enables controlled comparisons and reproducible alignment experiments.
}

\setkeywords{Large Language Models, Supervised Fine-Tuning, RLHF, Preference Optimization, Reward Modeling, Evaluation, Post-Training LLM}

\runningtitle{AlignTune: Modular Toolkit for Post-Training Alignment of Large Language Models}

\begin{document}
\maketitle

\section{Introduction}
\label{sec:introduction}

Large language models (LLMs) achieve strong performance across many tasks, but raw pre-trained models are often misaligned with downstream objectives helpfulness, safety, domain constraints, or product requirements. Post-training alignment via SFT~\cite{radford2019language}, preference optimization \cite{rafailov2023direct}, and RLHF~\cite{christiano2017deep,ouyang2022training} is therefore a necessary step for real-world deployment.

Practitioners today face a fragmented ecosystem. Most codebases target a single algorithm or backend, lack robust error handling, and resist integration into production pipelines. Libraries like TRL~\cite{trl2024} provide useful building blocks but expose one backend and a subset of algorithms, leaving users to assemble reward functions, evaluation, and configuration logic themselves. This fragmentation raises engineering overhead and makes it hard to reproduce or fairly compare alignment methods.
We argue that backend interference, reward fragmentation, and irreproducible pipelines are first-order obstacles in alignment research. Without standardized infrastructure, it is difficult to distinguish genuine methodological advances from implementation artifacts.

\emph{AlignTune} unifies supervised fine-tuning and RLHF-style training behind a single interface targeting multiple backends (TRL and Unsloth). A backend factory routes training requests to backend-specific implementations via a common API, while an environment-based isolation mechanism prevents Unsloth from globally patching \texttt{transformers} when TRL is selected. A unified configuration system supports reproducible experiments. AlignTune also provides an extensible reward framework (including domain-oriented reward functions for medical, legal, and financial settings), a reward-model training workflow, evaluation integration for standard benchmarks and custom tasks, and a CLI for end-to-end workflows. Table~\ref{tab:algo-support} summarizes supported algorithms; Section~\ref{sec:library-structure} gives architectural details.

\begin{figure}[t]
    \centering
    \includegraphics[width=0.63\textwidth]{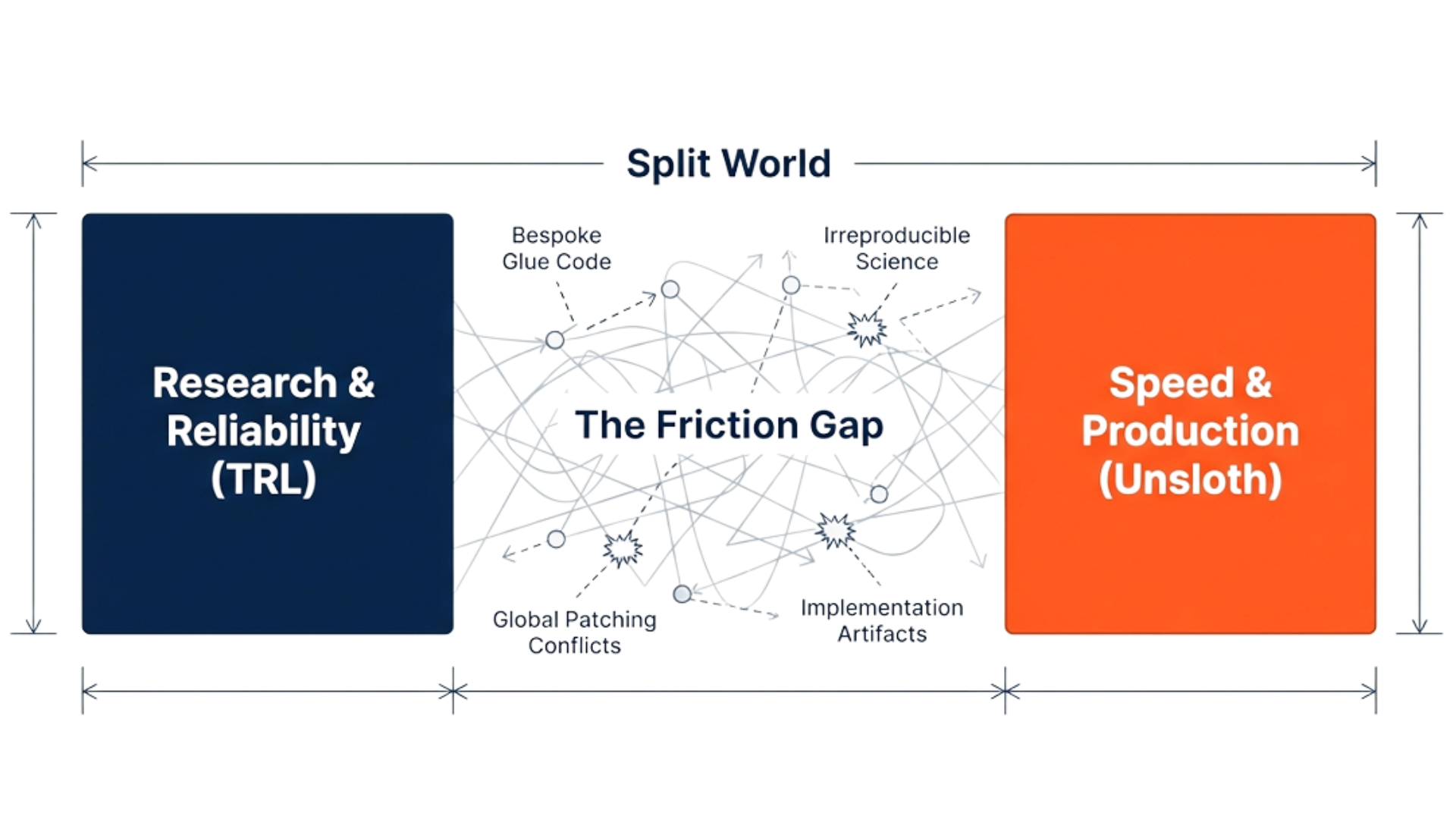}
    \vspace{-15pt}
    \caption{The current alignment ecosystem forces a choice between reliability and speed, fragmenting workflows and hindering reproducibility.}

    \label{fig:intro-fragmentation}
\end{figure}

\textbf{Contributions.}
This paper makes the following contributions:
\begin{itemize}
    \item A modular toolkit, AlignTune, that unifies SFT and RLHF-style training behind a single interface across TRL and Unsloth backends.
    \item A backend isolation mechanism that prevents Unsloth from patching \texttt{transformers} during pure TRL runs, with experimental validation (Section~\ref{subsec:isolation-test}).
    \item Backend benchmarks comparing TRL and Unsloth~\cite{unsloth2024} on throughput, memory, and evaluation metrics, showing backend-agnostic training without code changes (Section~\ref{subsec:backend-benchmark}).
    \item An extensible reward framework with 43 built-in reward functions, domain-specific signals, composable reward APIs, and a reward-model training pipeline (Section~\ref{subsec:reward-system}).
    \item A data management layer supporting Hugging Face Hub, JSON, CSV, Parquet, and directory-based sources.
\end{itemize}

\section{Installation and Getting Started}
\label{sec:installation}

AlignTune is distributed as a Python package and can be installed from PyPI or from source:

\begin{lstlisting}[language=bash,label={lst:install}]
git clone https://github.com/Lexsi-Labs/aligntune.git
cd aligntune
pip install -e .
\end{lstlisting}

After installation, verify the setup with:

\begin{lstlisting}[language=bash]
aligntune info      # prints environment, backend, and GPU details
\end{lstlisting}

A minimal training run requires only three lines of Python (here using the Alpaca~\cite{taori2023alpaca} dataset):

\begin{lstlisting}[language=Python,label={lst:minimal-sft}]
from aligntune.core.backend_factory import create_sft_trainer
trainer = create_sft_trainer(
    model_name="Qwen/Qwen2.5-0.5B-Instruct",
    dataset_name="tatsu-lab/alpaca",
    backend="trl", num_epochs=1, batch_size=4, learning_rate=5e-5)
trainer.train()
\end{lstlisting}

\section{Library Structure}
\label{sec:library-structure}

\subsection{Scope and Definitions}
\label{subsec:scope-defs}
\begin{figure}[t]
    \centering
    \includegraphics[width=0.99\textwidth]{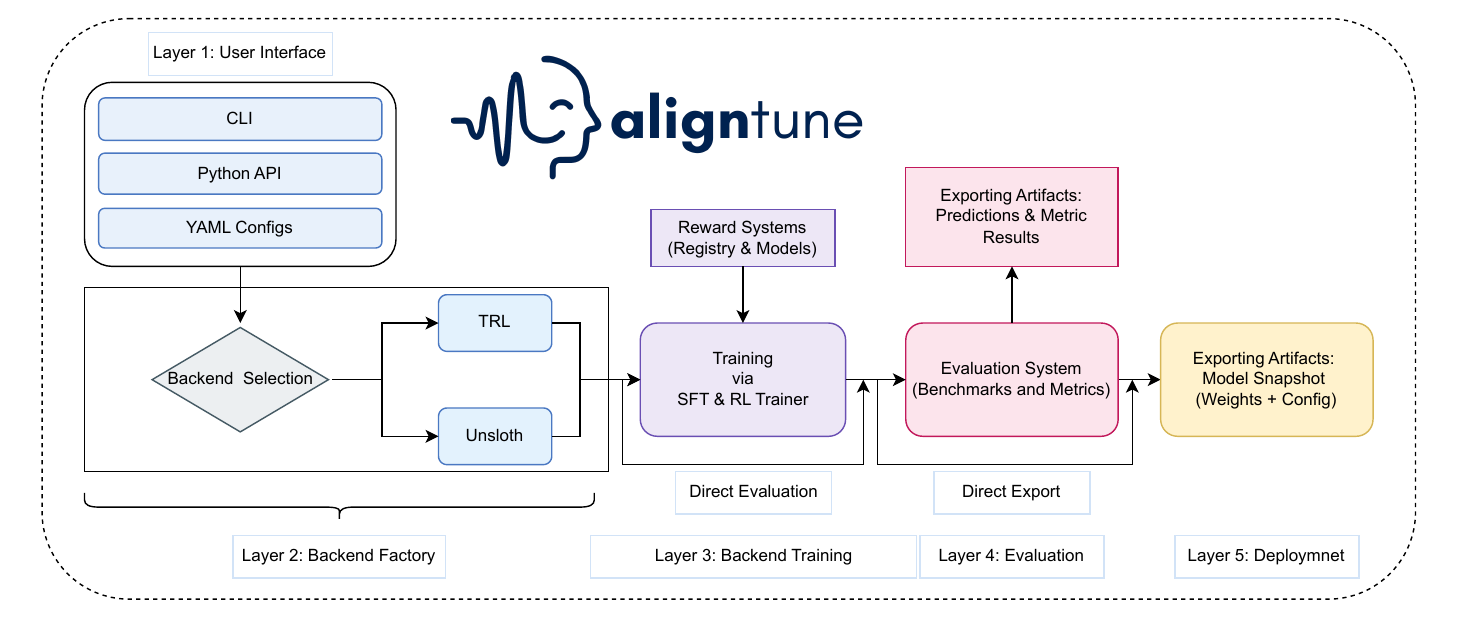}
    \caption{High-level architecture of AlignTune. Users interact via CLI, Python APIs, or YAML configs.
    The backend factory routes to TRL or Unsloth~\cite{unsloth2024} backends, which expose SFT and RL trainers. Reward and evaluation
    systems are shared across backends.}
    \label{fig:architecture-overview}
\end{figure}
Before describing the architecture, we clarify scope and define key terms. \emph{AlignTune} is the library presented in this work. A \emph{backend} is a concrete implementation of the training stack (e.g., TRL-based or Unsloth-accelerated).

The \emph{backend factory} is the component that selects and instantiates the appropriate backend from user configuration. The \emph{reward model pipeline} refers to the workflow of deriving training signals from rule-based reward functions, using them to train neural reward models, and then deploying those models within RLHF-style optimization.

\textbf{Scope : }
AlignTune supports: (1) SFT for instruction following and related supervised objectives;
(2) preference optimization (DPO~\cite{rafailov2023direct} and variants);
(3) policy optimization (PPO~\cite{schulman2017proximal}, GRPO~\cite{zhang2024group}, GSPO~\cite{zhang2024groupsequential}, and extensions);
(4) reward function composition and neural reward model training;
(5) evaluation via lm-eval and internal harnesses;
(6) YAML and Python API configuration; and
(7) a CLI for end-to-end workflows.

\textbf{Non-goals : }
AlignTune does not claim: (1) novel RLHF algorithms it standardizes existing methods;
(2) universal speedups acceleration depends on GPU and kernel compatibility;
(3) perfect feature parity between backends (see Table~\ref{tab:algo-support}); or
(4) support for all model families we target transformer-based LLMs compatible with Hugging Face Transformers~\cite{transformers2024}.

\subsection{High-Level Architecture}
\label{subsec:high-level-arch}

AlignTune organizes the alignment stack into layers: user interfaces (CLI, Python API, YAML configs), a backend factory, backend-specific trainers, and shared reward and evaluation systems. The core package is organized as:
\begin{itemize}
    \item \texttt{backends/}: backend implementations for TRL and Unsloth, each with \texttt{sft/} and \texttt{rl/} submodules.
    \item \texttt{core/}: shared core functionality, including the backend factory, RL and SFT configuration classes, 
    trainer bases, and registries.
    \item \texttt{rewards/}: reward function registry, reward-model training utilities, and reward-related types.
    \item \texttt{eval/}: evaluation framework, including lm-eval~\cite{lmeval2024} integration and custom tasks.
    \item \texttt{data/}: dataset processor and manager entry points.
    \item \texttt{cli/} and \texttt{cli\_commands/}: command-line entry points and configuration builders.
    \item \texttt{utils/}: supporting modules for device management, diagnostics, logging, validation, and model loading.
\end{itemize}

The design emphasizes modularity (each concern in a dedicated module), extensibility (new backends, algorithms, rewards, and tasks register without modifying core logic), and production readiness (error handling, diagnostics, typed configuration).

\subsection{Core API and Class Hierarchy}
\label{subsec:core-api}

AlignTune exposes a layered class hierarchy that users interact with at multiple levels of abstraction.

\subsubsection{Factory API}
The primary entry points are two factory functions in \texttt{aligntune.core.backend\_factory}:
\begin{itemize}
    \item \texttt{create\_sft\_trainer(\ldots)}   creates an SFT trainer for the requested backend and task type.
    \item \texttt{create\_rl\_trainer(\ldots)}   creates an RL trainer for the requested backend and algorithm.
\end{itemize}
Both functions accept a uniform set of keyword arguments (model name, dataset, backend, hyperparameters) and return a
trainer instance whose \texttt{.train()}, \texttt{.evaluate()}, and \texttt{.save\_model()} methods follow a common
protocol regardless of backend. Internally, the \texttt{BackendFactory} class dispatches to the correct
backend-specific trainer using enums: \texttt{TrainingType} (SFT or RL), \texttt{BackendType} (TRL or Unsloth), and
\texttt{RLAlgorithm} (DPO, PPO, GRPO, GSPO, DAPO, Dr.~GRPO, GBMPO, Counterfactual GRPO, BOLT, GBMPO). A \texttt{BackendConfig} dataclass stores backend selection, isolation flags, and fallback preferences.

\subsubsection{Trainer Class Hierarchy}
All trainers derive from one of two abstract base classes:
\begin{itemize}
    \item \texttt{TrainerBase} (RL): defines the full lifecycle for reinforcement learning style training, including reward integration, rollout generation, policy updates, and checkpoint management. It maintains a \texttt{TrainingState} dataclass that tracks the current step and epoch, the best observed metric, and the active checkpoint path.
    \item \texttt{SFTTrainerBase} (SFT): defines the lifecycle for supervised fine-tuning, including task-aware data preparation, orchestration of the training loop, and evaluation hooks.
\end{itemize}

\noindent Each backend provides concrete trainer implementations.

\textbf{TRL backend} : \texttt{TRLSFTTrainer}, \texttt{TRLDPOTrainer}, \texttt{TRLPPOTrainer},
\texttt{TRLGRPOTrainer}, \texttt{TRLGSPOTrainer}, \texttt{TRLDAPOTrainer}, \texttt{TRLDRGRPOTrainer},
\texttt{TRLGBMPOTrainer}, \texttt{TRLCounterFactGRPOTrainer}, \texttt{TRLPACETrainer}.

\textbf{Unsloth backend} : mirrors the TRL set using an \texttt{Unsloth} prefix (e.g., \texttt{UnslothDPOTrainer}). GSPO, GBMPO, and Meta-ES are currently TRL-only (see Table~\ref{tab:algo-support}).

\noindent For supervised fine-tuning, an \texttt{SFTTrainerFactory} additionally dispatches requests by \texttt{TaskType} (instruction following, text classification, token classification, text generation, or chat completion). Each task can use specialised data formatting and loss computation (e.g., a \texttt{ClassificationTrainer} for classification objectives).

\begin{figure}[t]
    \centering
    \includegraphics[width=0.63\textwidth]{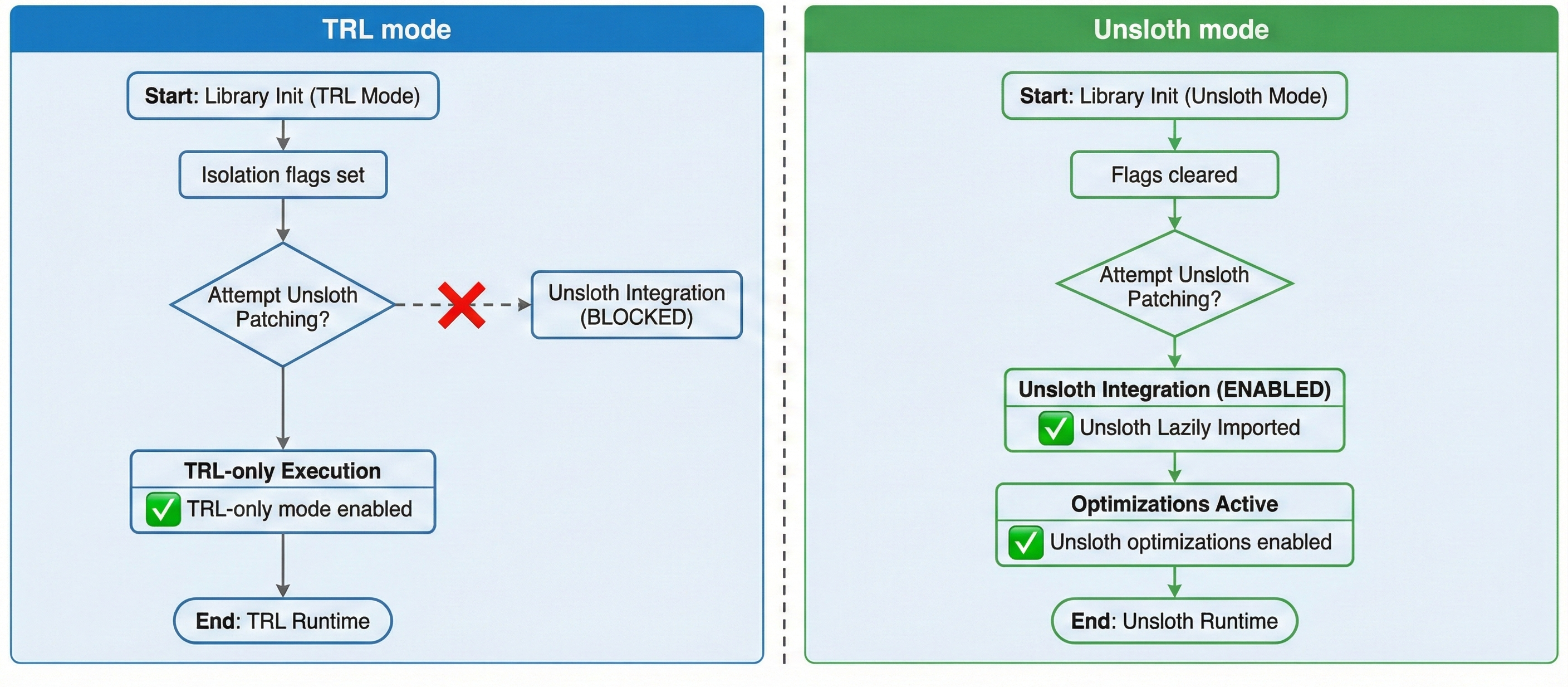}
    \caption{Backend isolation flow. TRL runs block Unsloth patches; Unsloth runs clear isolation flags and import lazily.}
    \label{fig:backend-isolation}
\end{figure}
\subsection{Multi-Backend Architecture}
\label{subsec:multi-backend}

AlignTune's multi-backend architecture lets users choose between:
\begin{itemize}
    \item TRL: pure TRL implementations optimized for reliability and compatibility.
    \item Unsloth: accelerated implementations offering speed-ups and memory savings via quantization and optimized kernels.
\end{itemize}

The backend factory centralizes selection and instantiation. Without it, users must manually configure backend-specific parameters differently for each backend, creating configuration skew that conflates backend effects with setup differences. 

The following examples use DialoGPT~\cite{zhang2019dialogpt} with Alpaca~\cite{taori2023alpaca} for SFT and HH-RLHF~\cite{anthropic2023hh} for DPO:

\begin{lstlisting}[language=Python,label={lst:backend-factory-sft}]
from aligntune.core.backend_factory import create_sft_trainer

# SFT trainer (pure TRL backend)
sft_trainer = create_sft_trainer(
    model_name="microsoft/DialoGPT-medium",
    dataset_name="tatsu-lab/alpaca",
    backend="trl",
    num_epochs=1,
    batch_size=4,
    learning_rate=5e-5,
)
\end{lstlisting}
\begin{lstlisting}[language=Python,caption={DPO trainer via backend factory},label={lst:backend-factory-dpo}]
from aligntune.core.backend_factory import create_rl_trainer
# DPO trainer (Unsloth backend)
dpo_trainer = create_rl_trainer(
    model_name="unsloth/Llama-3.2-1B-Instruct-bnb-4bit",
    dataset_name="Anthropic/hh-rlhf",
    algorithm="dpo",
    backend="unsloth",
    num_epochs=1,
    batch_size=1,
    learning_rate=1e-6,
)
\end{lstlisting}

\subsubsection{Backend Isolation System}

Unsloth patches the transformers stack globally to insert optimized kernels and compression logic (Figure~\ref{fig:backend-isolation}). This is desirable when Unsloth~\cite{unsloth2024} is explicitly selected, but can interfere with pure TRL runs in the same environment. Without isolation, identical configs with different import order can produce different training dynamics, invalidating backend comparisons. AlignTune's isolation system has four components:
\begin{itemize}
    \item \emph{Environment-variable control.} When TRL~\cite{trl2024} is selected, the factory sets \texttt{PURE\_TRL\_MODE}, \texttt{TRL\_ONLY\_MODE}, and \texttt{DISABLE\_UNSLOTH\_FOR\_TRL} to block Unsloth patches. When Unsloth is requested, these flags are cleared and Unsloth is imported lazily.
    \item \emph{Lazy loading.} A helper in \texttt{\_imports.py} checks Unsloth availability (PyTorch, CUDA versions) but defers the actual import until needed.
    \item \emph{String-based selection.} The factory accepts backend names as strings, avoiding enum imports that could trigger Unsloth initialization.
    \item \emph{Automatic fallback.} If Unsloth is unavailable or fails compatibility checks, informative errors point users to TRL~\cite{trl2024} as a fallback.
\end{itemize}

\subsection{Training Algorithms}
\label{subsec:training-algorithms}

AlignTune exposes both supervised and reinforcement-learning-based alignment methods under a unified interface.

\textbf{Supervised Fine-Tuning (SFT) : }
SFT covers instruction following, text classification, token classification, and chat-style tasks. The SFT stack supports:
(i) task routing configuration-level selection of task type, mapping to the appropriate trainer;
(ii) parameter-efficient fine-tuning via LoRA~\cite{hu2021lora} and QLoRA~\cite{dettmers2023qlora}, including 4-bit quantization; and
(iii) gradient checkpointing, mixed precision (fp16/bf16), and dataset packing.
SFT trainers are available for both TRL and Unsloth backends with a common configuration surface.

\textbf{Reinforcement Learning and RLHF Algorithms : }
The goal is not to introduce new algorithms, but to show that a single abstraction can host a broad class of alignment methods without backend-specific rewrites.

\begin{table}[pt]
    \centering
    \footnotesize
    \caption{Algorithm support matrix in AlignTune}
    \label{tab:algo-support}
    \begin{tabular}{lccc}
        \toprule
        Algorithm & TRL Backend & Unsloth Backend & Description \\
        \midrule
        SFT~\cite{radford2019language} & \checkmark & \checkmark & Supervised instruction/task fine-tuning \\
        DPO~\cite{rafailov2023direct} & \checkmark & \checkmark & Direct Preference Optimization (no reward model) \\
        PPO~\cite{schulman2017proximal} & \checkmark & \checkmark & Proximal Policy Optimization with explicit rewards \\
        GRPO~\cite{zhang2024group} & \checkmark & \checkmark & Group Relative Policy Optimization \\
        GSPO~\cite{zhang2024groupsequential} & \checkmark & \checkmark & Group Sequential Policy Optimization \\
        DAPO~\cite{yu2025dapo} & \checkmark & \checkmark & Decoupled variant addressing GRPO limitations \\
        Dr.\ GRPO~\cite{liu2025drgrpo} & \checkmark & \checkmark & Unbiased GRPO variant (length-bias correction) \\
        GBMPO~\cite{yuan2025gbmpo} & \checkmark & \texttimes & Group-Based Mirror Policy Optimization \\
        Counterfactual GRPO~\cite{khandoga2026uniformcreditcausalcredit} & \checkmark & \checkmark & Counterfactual extension of GRPO \\
        PACE~\cite{khandoga2025bolt} & \checkmark & \checkmark & Curriculum Enhancement for Sample-Efficient GRPO \\
        \bottomrule
    \end{tabular}
\end{table}
For each algorithm, AlignTune implements both TRL-based and, where applicable, Unsloth-based trainers that share a
common base class. Table~\ref{tab:algo-support} summarizes the supported RLHF algorithms and backend coverage. PPO trainers additionally support:
(i) reward model integration from the Hugging Face Hub, local checkpoints, or AlignTune's own reward-model pipeline;
(ii) model-family consistency checks (e.g., ensuring compatible policy and reward model families such as Qwen~\cite{qwen2024}, LLaMA~\cite{touvron2023llama}, Mistral~\cite{jiang2023mistral}); and
(iii) explicit KL penalty control, clipping settings, and multi-task reward configurations.

\subsection{Reward System}
\label{subsec:reward-system}
\begin{table}[pt]
    \centering
    \caption{Catalog of built-in reward functions in AlignTune, grouped by category. All reward functions can be combined with configurable weights to form composite rewards. Custom reward functions can be registered via the \texttt{RewardRegistry} API.}
    \label{tab:reward-catalog}
    \small
    \begin{tabular}{lp{9cm}}
        \toprule
        Category & Reward Functions \\
        \midrule
        Basic Quality &
        length, coherence \\
        \addlinespace
        Task-Specific &
        sentiment, safety, factuality, bias \\
        \addlinespace
        Code Quality &
        code\_syntax, code\_execution, code\_completeness, code\_quality, code\_correctness \\
        \addlinespace
        Math \& Reasoning &
        math\_correctness, logical\_consistency, commonsense, math\_reasoning, counterfactual\_math \\
        \addlinespace
        Specialized Alignment &
        hallucination, toxicity, politeness, helpfulness, honesty \\
        \addlinespace
        Enhanced Quality &
        diversity, fluency, relevance, brevity \\
        \addlinespace
        Instruction \& Alignment &
        instruction\_following, harmlessness, conciseness \\
        \addlinespace
        Context \& Temporal &
        context\_relevance, temporal\_consistency \\
        \addlinespace
        Advanced Metrics &
        semantic\_similarity, readability, engagement \\
        \addlinespace
        Domain-Specific &
        medical\_accuracy, legal\_compliance, financial\_accuracy \\
        \addlinespace
        Advanced Reasoning &
        causal\_reasoning, counterfactual\_reasoning \\
        \addlinespace
        Benchmark-Specific &
        mbpp\_reward~\cite{austin2021program} (code generation benchmark) \\
        \bottomrule
    \end{tabular}
\end{table}

\subsubsection{Reward Class Hierarchy}

The reward subsystem is built on a layered class hierarchy. \texttt{RewardFunction} is the abstract base class; every
reward implements a \texttt{compute(text, **kwargs) -> float} method. A \texttt{RewardType} enum (30+ members)
categorises each function, and a \texttt{RewardConfig} dataclass stores parameters such as weight, threshold, and
normalization mode.

Key infrastructure classes:
\begin{itemize}
    \item \texttt{RewardFunctionFactory}   creates reward function instances from string keys or
    \texttt{RewardType} values.
    \item \texttt{CompositeReward}   combines multiple \texttt{RewardFunction} instances with configurable weights,
    enabling multi-objective reward signals (e.g., $0.3\times\text{length} + 0.4\times\text{sentiment} +
    0.3\times\text{safety}$).
    \item \texttt{RewardRegistry}   central registry that maps string keys to reward types, manages default
    configurations, and exposes \texttt{register\_custom\_reward()} and \texttt{get\_reward\_function()} helpers.
\end{itemize}

AlignTune ships concrete reward classes implementing the \texttt{RewardFunction} interface, organised into
multiple categories (see Table~\ref{tab:reward-catalog}). Notable implementations include
\texttt{CodeExecutionReward} (sandboxed code execution with test-case validation),
\texttt{MathCorrectnessReward} (symbolic and numeric answer grading),
(domain-specific scoring), and \texttt{MBPPReward} (code generation benchmark reward).

Without a centralised registry, reward logic scatters across trainer implementations, making it hard to audit which rewards apply in which experiments and leading to inconsistent application across runs.

\begin{figure}[t]
    \centering
    \includegraphics[width=0.72\linewidth]{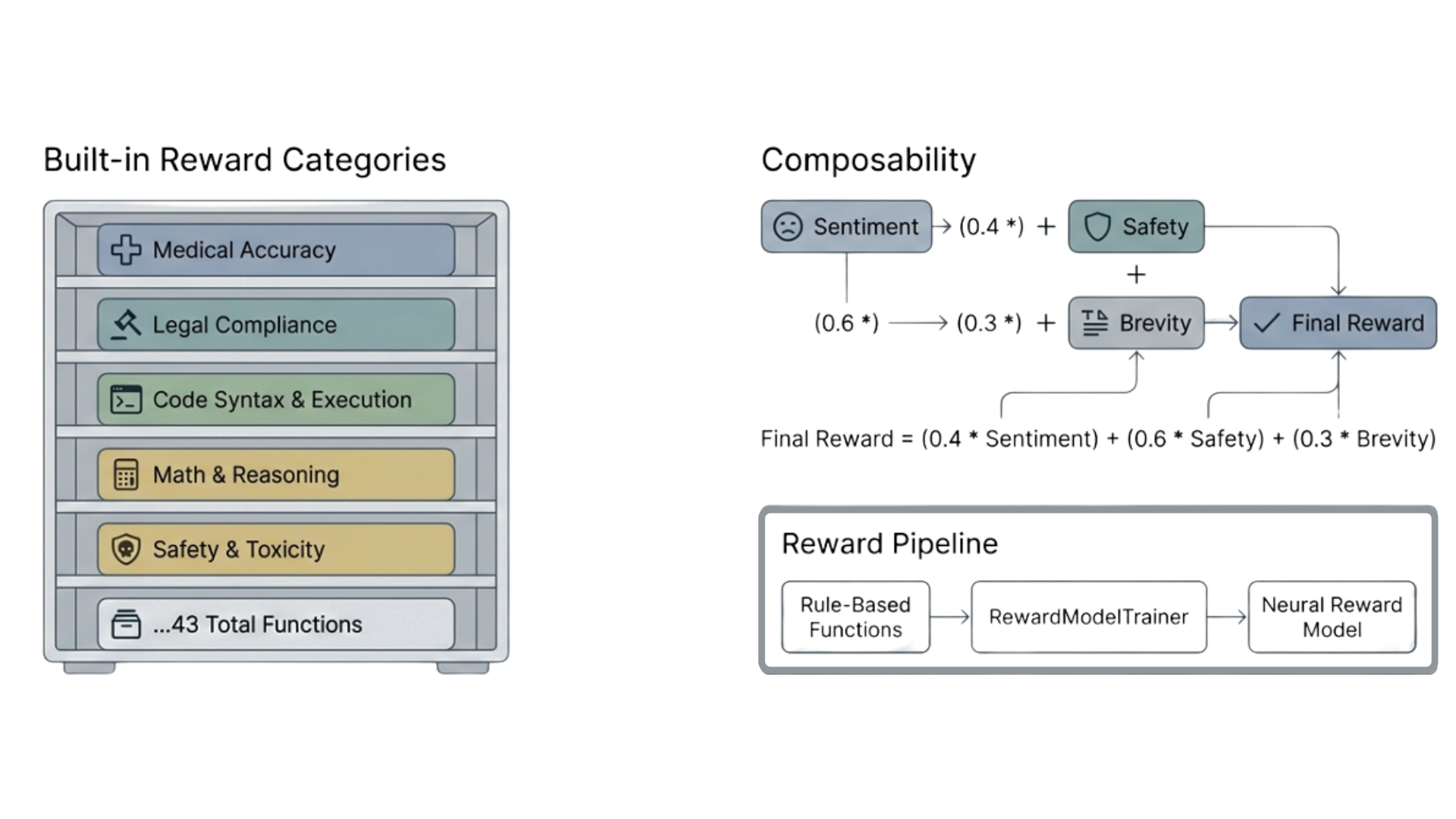}
    \vspace{-20pt}
    \caption{AlignTune reward system: built-in reward categories, weighted composition, and the pipeline from rule-based rewards to neural reward models.}
    \label{fig:reward-system-overview}
\end{figure}

\subsubsection{Reward Function Registry}

The registry maps string keys to reward types, manages default configurations, and provides helpers to construct
composite rewards.

New reward functions can be registered as follows:

\begin{lstlisting}[language=Python,label={lst:custom-reward}]
from aligntune.rewards.core import RewardFunction, RewardConfig, RewardType
from aligntune.rewards.registry import RewardRegistry

class CustomReward(RewardFunction):
    def __init__(self):
        super().__init__(RewardType.CUSTOM)

    def compute(self, text: str, **kwargs) -> float:
        # Custom scoring logic
        return 0.8

# Register the reward for use in configs
RewardRegistry.register_custom_reward("custom", CustomReward)
\end{lstlisting}

\subsubsection{Reward-Model Training Pipeline}
Beyond rule-based rewards, AlignTune supports training neural~\cite{christiano2017deep,ouyang2022training} reward models from text data labeled by reward
functions. The pipeline is implemented by four classes in \texttt{aligntune.rewards.training}:
\begin{itemize}
    \item \texttt{RewardModelTrainer}   orchestrates end-to-end reward model training: generating labeled data from
    rule-based functions, training a transformer-based reward model, and saving checkpoints.
    \item \texttt{RewardModelDataset}   a PyTorch \texttt{Dataset} that pairs texts with composite reward scores.
    \item \texttt{RewardModelValidator}   evaluates reward model accuracy, calibration, and correlation with
    ground-truth reward functions.
    \item \texttt{RewardModelLoader}   loads trained reward models for inference, including
    \texttt{TRLCompatibleRewardModel} for direct integration with TRL PPO trainers.
\end{itemize}
The typical workflow is:
\begin{enumerate}
    \item Choose a base model architecture for the reward model.
    \item Define a set of reward functions and associated weights.
    \item Generate training examples and compute composite rewards.
    \item Train the reward model via the reward training module.
    \item Plug the resulting reward model into PPO (or other RL trainers).
\end{enumerate}

By treating rewards as first-class objects, AlignTune also enables controlled experiments over reward structure (e.g., sparse vs.\ dense, rule-based vs.\ learned) something difficult when reward logic is coupled to specific trainers. This supports reward ablations, audits, and systematic studies of how reward design affects alignment outcomes.

\begin{lstlisting}[language=Python,caption={Reward model training and PPO integration},label={lst:reward-training}]
from aligntune.rewards.training import RewardModelTrainer
from aligntune.rewards.registry import RewardRegistry
from aligntune.core.backend_factory import create_rl_trainer

registry = RewardRegistry()
length_func = registry.get_reward_function("length")
sentiment_func = registry.get_reward_function("sentiment")
safety_func = registry.get_reward_function("safety")

# 1. Train a neural reward model from rule-based rewards
rm_trainer = RewardModelTrainer(
    base_model_name="microsoft/DialoGPT-medium",
    reward_functions=[length_func, sentiment_func, safety_func],
    composite_weights=[0.3, 0.4, 0.3],
)

training_data = rm_trainer.generate_training_data(
    texts=training_texts_list,
    batch_size=2  
)

reward_model_path = rm_trainer.train_reward_model(
    training_data=training_data,
    output_dir="./reward_models/custom",
    num_epochs=3,
    learning_rate=1e-5,
    batch_size=4 
)

# 2. Use the trained reward model in PPO
ppo_trainer = create_rl_trainer(
    model_name="TinyLlama/TinyLlama-1.1B-Chat-v1.0",
    dataset_name="Anthropic/hh-rlhf",
    algorithm="ppo",
    backend="unsloth",
    reward_model_path=reward_model_path,
    num_epochs=1,
    batch_size=1,
    learning_rate=2e-4,
)
ppo_trainer.train()
\end{lstlisting}

\subsection{Data Management}
\label{subsec:data-management}

AlignTune provides a unified data management layer in \texttt{aligntune.data} that abstracts over heterogeneous data
sources. The  \texttt{DataManager} class coordinates loading, processing, and caching of training and
evaluation datasets.

\textbf{Loaders : }
A \texttt{LoaderResolver} inspects the data source string and dispatches to the appropriate loader. Several concrete loaders extend the \texttt{BaseLoader} interface:
\begin{itemize}
    \item \texttt{HFLoader}   loads datasets from the Hugging Face Hub via the \texttt{datasets} library, with
    support for streaming, splits, and column selection.
    \item \texttt{JSONLoader}, \texttt{CSVLoader}, \texttt{ParquetLoader}   load from local files in the
    corresponding formats.
    \item \texttt{DirectoryLoader}   loads from a local directory of text or structured files, with configurable
    file-type filtering.
\end{itemize}
All loaders return a common \texttt{Dataset} object, ensuring that downstream trainers and evaluators are agnostic
to the data origin. A \texttt{DatasetCache} accelerates repeated experiments by caching processed datasets.

\subsection{Configuration and CLI}
\label{subsec:config-cli}

\textbf{Unified Configuration System : }
AlignTune's configuration system is centered on strongly-typed dataclasses:
\begin{itemize}
    \item \texttt{RLConfig} for RL training, with nested sections for \texttt{algo}, \texttt{model}, 
    \texttt{datasets}, \texttt{train}, \texttt{logging}, \texttt{rewards}, and \texttt{caching}.
    \item \texttt{SFTConfig} for SFT, with analogous model, dataset, train, and logging sections.
\end{itemize}

Configurations can be authored in YAML, separating code from hyperparameters. Validation logic provides informative errors and estimates memory usage to catch misconfigurations early.

\textbf{Command-Line Interface and Recipes.}
The \texttt{aligntune} CLI offers high-level commands:
\begin{itemize}
    \item \texttt{aligntune info}: environment and backend information.
    \item \texttt{aligntune train}: run SFT or RL training from a config or inline arguments.
    \item \texttt{aligntune diagnose}: run environment diagnostics.
    \item \texttt{aligntune recipes}: list, show, and copy pre-defined recipes.
\end{itemize}

Example usage:

\begin{lstlisting}[language=bash,caption={CLI usage with YAML configs},label={lst:cli-usage}]
aligntune train --config examples/grpo_gsm8k_trl/config_grpo_gsm8k.yaml
\end{lstlisting}

\begin{lstlisting}[language=python,caption={YAML configuration sketch for a DPO experiment},label={lst:dpo-yaml}]
algo: dpo

model:
  name_or_path: "Qwen/Qwen3-0.6B"
  backend: trl
  max_seq_length: 512

datasets:
  - name: "Anthropic/hh-rlhf"
    split: "train"
    max_samples: 100

train:
  max_steps: 100
  per_device_batch_size: 4
  learning_rate: 0.00005
  gradient_accumulation_steps: 1
  beta: 0.1
  max_prompt_length: 512
  max_completion_length: 512

logging:
  output_dir: "./output/dpo_qwen"
  log_level: "INFO"
  save_steps: 100
  eval_steps: 100

chat_template: auto
\end{lstlisting}

Recipes encode best-practice configurations for families such as LLaMA~\cite{touvron2023llama} and Qwen~\cite{qwen2024}, handling authentication and 
model-specific quirks.

\subsection{Evaluation System}
\label{subsec:evaluation-system}

The evaluation subsystem integrates both standardised benchmarks and custom tasks through a class hierarchy rooted
in \texttt{aligntune.eval}.

\begin{figure}[t]
    \centering
    \includegraphics[width=0.54\linewidth]{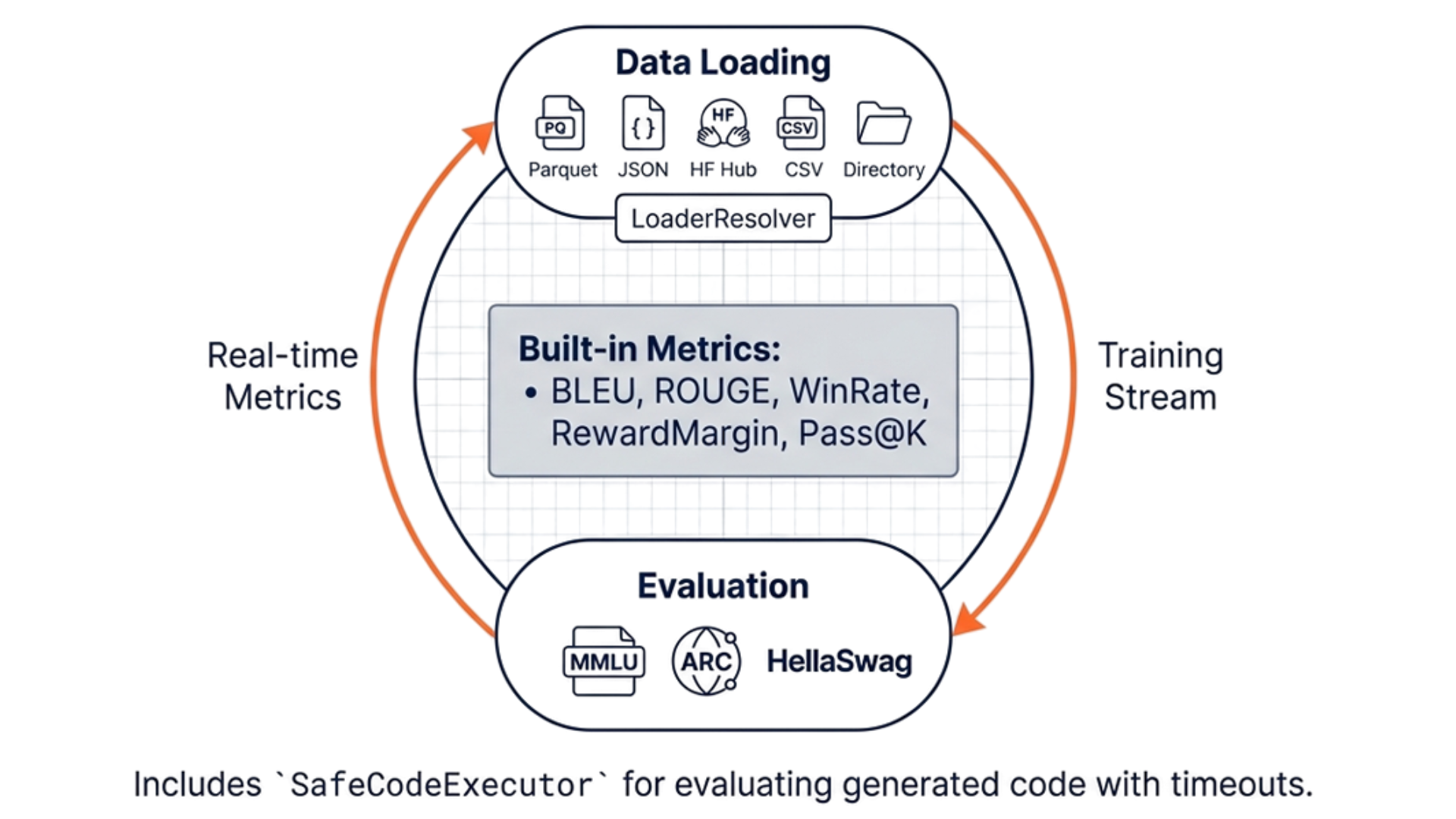}
    \caption{Evaluation and monitoring pipeline: data loading, real-time metrics, benchmark evaluation, and sandboxed code execution.}
    \label{fig:evaluation-pipeline}
\end{figure}

\subsubsection{Evaluation Infrastructure}
\begin{itemize}
    \item \texttt{BaseEvaluator} and \texttt{RLEvaluator}   abstract base and RL-specific evaluators that orchestrate
    metric computation over model outputs.
    \item \texttt{EvalConfig}   typed configuration specifying tasks, metrics, sample sizes, and logging options.
    \item \texttt{EvalTask} and \texttt{EvalResult}   represent individual evaluation tasks and their results.
    \item \texttt{EvalRunner}   orchestrates multi-task evaluation runs, dispatching to registered metrics.
    \item \texttt{EvalRegistry}   registry of evaluation functions and tasks, enabling extensibility.
    \item \texttt{SafeCodeExecutor}   sandboxed code execution environment with timeout (\texttt{TimeoutException})
    for evaluating generated code safely.
\end{itemize}

 \subsubsection{In-Built Metrics }
AlignTune provides multiple metric implementations extending an abstract \texttt{Metric} base class:
Text metrics: \texttt{RougeMetric}, \texttt{BleuMetric}.
Generic metrics: \texttt{PerplexityMetric}, \texttt{AccuracyMetric}.
RL-specific metrics\textbf{:} \texttt{KLDivergenceMetric} (policy divergence from reference),
\texttt{RewardAccuracyMetric}, \texttt{PolicyEntropyMetric}.
DPO-specific metrics: \texttt{WinRateMetric}, \texttt{RewardMarginMetric},
\texttt{PreferenceAccuracyMetric}, \texttt{LogRatioMetric}, \texttt{ImplicitRewardMetric},
\texttt{CalibrationMetric}.
Specialised metrics\textbf{:} \texttt{PassAtKMetric} (code generation pass@$k$),
\texttt{MathAccuracyMetric}.

\subsubsection{Benchmark Integration}
\texttt{LMEvalRunner} and \texttt{LMEvalConfig} wrap the lm-eval-harness~\cite{lmeval2024} for standardised
benchmarks such as HellaSwag~\cite{zellers2019hellaswag}, ARC~\cite{clark2018arc}, and
MMLU~\cite{hendrycks2021measuring}. Custom evaluation tasks for text generation, classification, summarisation,
code, and math can be registered via the \texttt{EvalRegistry}.
\textsc{Real-time monitoring:} A \texttt{SampleLogger} periodically generates qualitative outputs (e.g., at 50\%
of training steps) to monitor regressions. Training and evaluation flows follow a simple pattern: models are loaded,
evaluation datasets are prepared, inference is run, and metrics are computed and logged.
\subsubsection{Evaluation Usage Examples}

AlignTune's evaluation system supports diverse workflows. Here we demonstrate standalone evaluation 
and integrated training and evaluation pipelines.

\textbf{Math Task Evaluation : }
For mathematical reasoning tasks, we can evaluate a trained model on GSM8K~\cite{cobbe2021trainingverifierssolvemath}:

\begin{lstlisting}[language=Python,caption={Evaluating math reasoning with GSM8K},label={lst:math-eval}]
from aligntune.eval.runner import EvalConfig, run_eval

eval_config = EvalConfig(
    model_path="./output/grpo_model",
    output_dir="./results/math_eval",
    task_type="math",
    data_task_type="grpo",
    dataset_name="openai/gsm8k",
    dataset_config="main",
    split="test",
    max_samples=100,
    batch_size=8,
    temperature=0.1,
    max_length=512,
    use_lora=True,
    base_model="meta-llama/Llama-3.2-3B-Instruct",
    use_unsloth=True,
    metrics=["math_accuracy"],
    column_mapping={"question": "prompt", "answer": "response"},
)

results = run_eval(eval_config)
\end{lstlisting}
\newpage
\textbf{Text Generation Evaluation.}
For instruction-following or dialogue tasks, we evaluate on standard text metrics:

\begin{lstlisting}[language=Python,caption={Text generation evaluation},label={lst:text-eval}]
from aligntune.eval.runner import EvalConfig, run_eval

text_eval_config = EvalConfig(
    model_path="Qwen/Qwen3-4B-Instruct-2507",
    device="cuda",
    use_unsloth=True,
    task_type="text",
    data_task_type="sft",
    metrics=["rouge", "bleu", "perplexity"],
    dataset_name="HuggingFaceH4/ultrachat_200k",
    split="test_sft",
    max_samples=100,
    max_length=1024,
    temperature=0.7,
    output_dir="./results/text_eval",
)

results = run_eval(text_eval_config)
\end{lstlisting}

These examples illustrate task-specific evaluation (math vs.\ text), adapter handling (\texttt{use\_lora}), 
and metric selection, all through a unified \texttt{EvalConfig} interface.
\begin{figure}[t]
    \centering
    \includegraphics[width=0.9\textwidth]{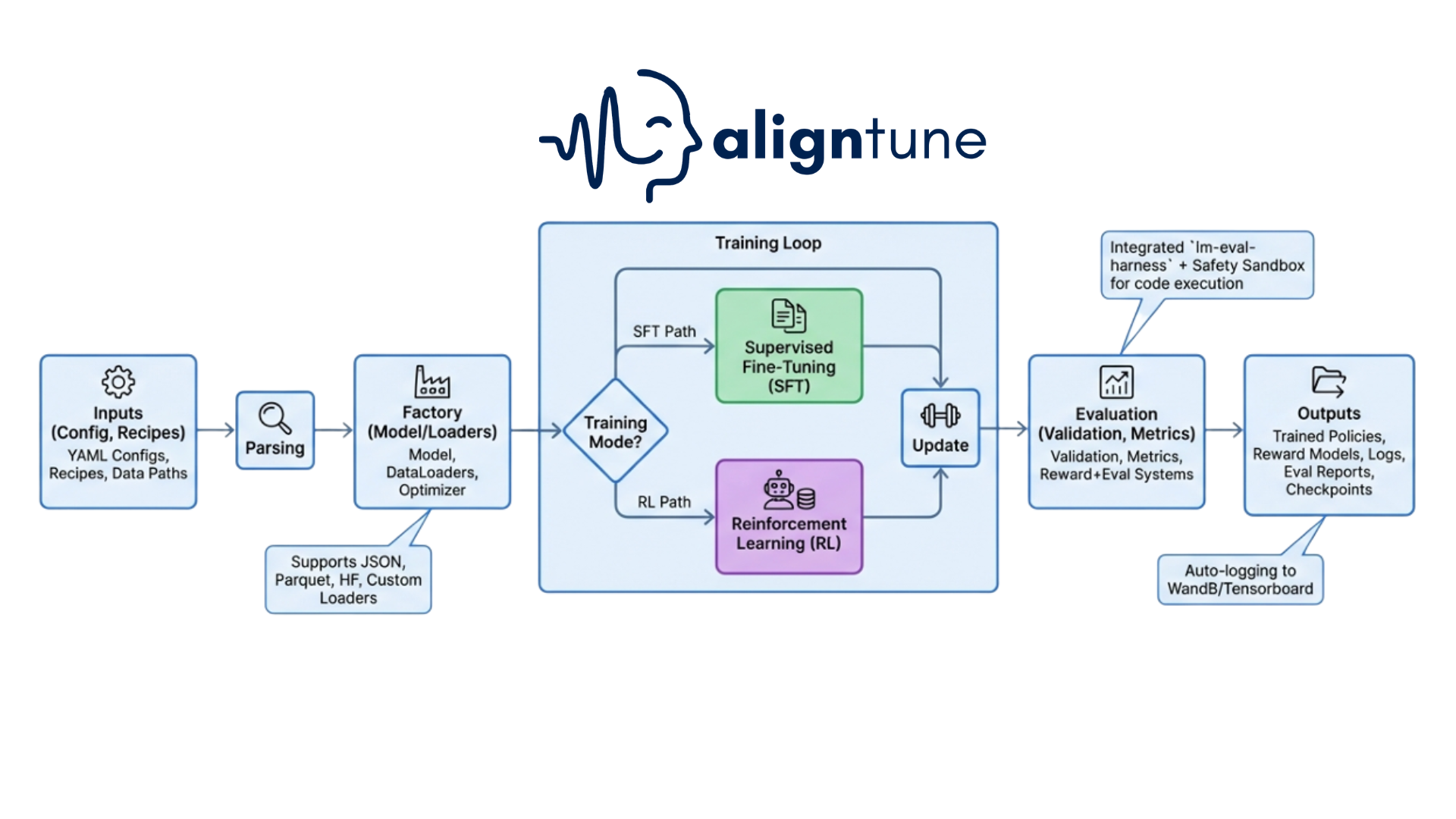}
    \vspace{-50pt}
    \caption{End-to-end alignment pipeline. Configurations and recipes feed the backend factory, which creates trainers that interact with reward and evaluation systems.}
    \label{fig:alignment-pipeline}
\end{figure}

\subsection{Utilities and Production Features}
\label{subsec:utilities}

AlignTune ships a comprehensive utility layer in \texttt{aligntune.utils} designed for production-grade training
workflows.

\textbf{Model and device management : }
\texttt{ModelLoader} handles quantisation (4-bit/8-bit via bitsandbytes~\cite{bitsandbytes2024}), LoRA~\cite{hu2021lora} adapter injection via PEFT~\cite{peft2024}, and automatic dtype selection. \texttt{DeviceManager} manages GPU/CPU allocation via Accelerate~\cite{accelerate2024}. \texttt{CheckpointManager} saves and resumes full training state.

\textbf{Error hierarchy:}
A structured hierarchy provides actionable diagnostics: \texttt{AlignTuneError} $\rightarrow$ \texttt{ConfigurationError}, \texttt{TrainingError}, \texttt{EnvironmentError}, \texttt{ValidationError}. Each error carries context and suggested fixes.

\textbf{Health monitoring : }
\texttt{HealthMonitor} tracks loss spikes, gradient norms, and memory pressure. \texttt{TrainingDiagnostics} and \texttt{TrainingMonitor} provide real-time metric dashboards. \texttt{DiagnosticsCollector} aggregates GPU utilisation, memory, and disk statistics.

\textbf{Configuration validation : }
\texttt{ConfigValidator} checks typed configs against schemas, validates required fields, estimates peak GPU memory, and warns about misconfigurations before training starts.

\section{Illustrative Examples}
\label{sec:examples-recipes}

A typical example follows this pattern:

\begin{lstlisting}[language=Python,caption={Direct-API GRPO example (abridged)},label={lst:grpo-example}]
from aligntune.core.backend_factory import create_rl_trainer

trainer = create_rl_trainer(
    model_name="Qwen/Qwen2.5-0.5B-Instruct",
    dataset_name="openai/gsm8k",
    config_name="main",
    algorithm="grpo",
    backend="trl",
    num_epochs=1,
    batch_size=4,
    learning_rate=5e-6,
    num_generations=4,
)

trainer.train()
metrics = trainer.evaluate()
\end{lstlisting}

AlignTune also supports end-to-end training and evaluation pipelines for supervised fine-tuning:

\begin{lstlisting}[language=Python,caption={End-to-end SFT and evaluation pipeline},label={lst:sft-eval-example}]
from aligntune.core.backend_factory import create_sft_trainer
from aligntune.eval.runner import EvalConfig, run_eval

# Train a domain-specific model
trainer = create_sft_trainer(
    model_name="Qwen/Qwen3-4B-Instruct-2507",
    dataset_name="sohamb37lexsi/Bitext-wealth-management-llm-chatbot-training-dataset",
    backend="trl",
    output_dir="./wealth_management_model",
    task_type="instruction_following",
    max_steps=500,
    epochs=50,
    batch_size=2,
    gradient_accumulation_steps=2,
    learning_rate=2e-4,
    warmup_ratio=0.1,
    max_seq_length=1512,
    dataset_text_field="messages",
    split="train",
    use_peft=True,
    lora_r=6,
    lora_alpha=8,
    lora_dropout=0.05,
    lora_target_modules="all-linear",
    quantization={
        "load_in_4bit": True,
        "bnb_4bit_use_double_quant": True,
        "bnb_4bit_quant_type": "nf4",
        "bnb_4bit_compute_dtype": "bfloat16",
    },
    bf16=True,
    tf32=True,
    trust_remote_code=True,
    seed=42,
)

trainer.train()

# Evaluate the trained model
eval_cfg = EvalConfig(
    model_path="./wealth_management_model",
    output_dir="./results/wealth_management",
    data_task_type="sft",
    dataset_name="sohamb37lexsi/Bitext-wealth-management-llm-chatbot-training-dataset",
    split="test",
    device="cuda:0",
    batch_size=16,
    metrics=["bleu", "rouge"],
)

results = run_eval(eval_cfg)
\end{lstlisting}

\section{Case Studies}
\label{sec:case-studies}
\begin{figure}[pt]
    \centering
    \includegraphics[width=0.45\linewidth]{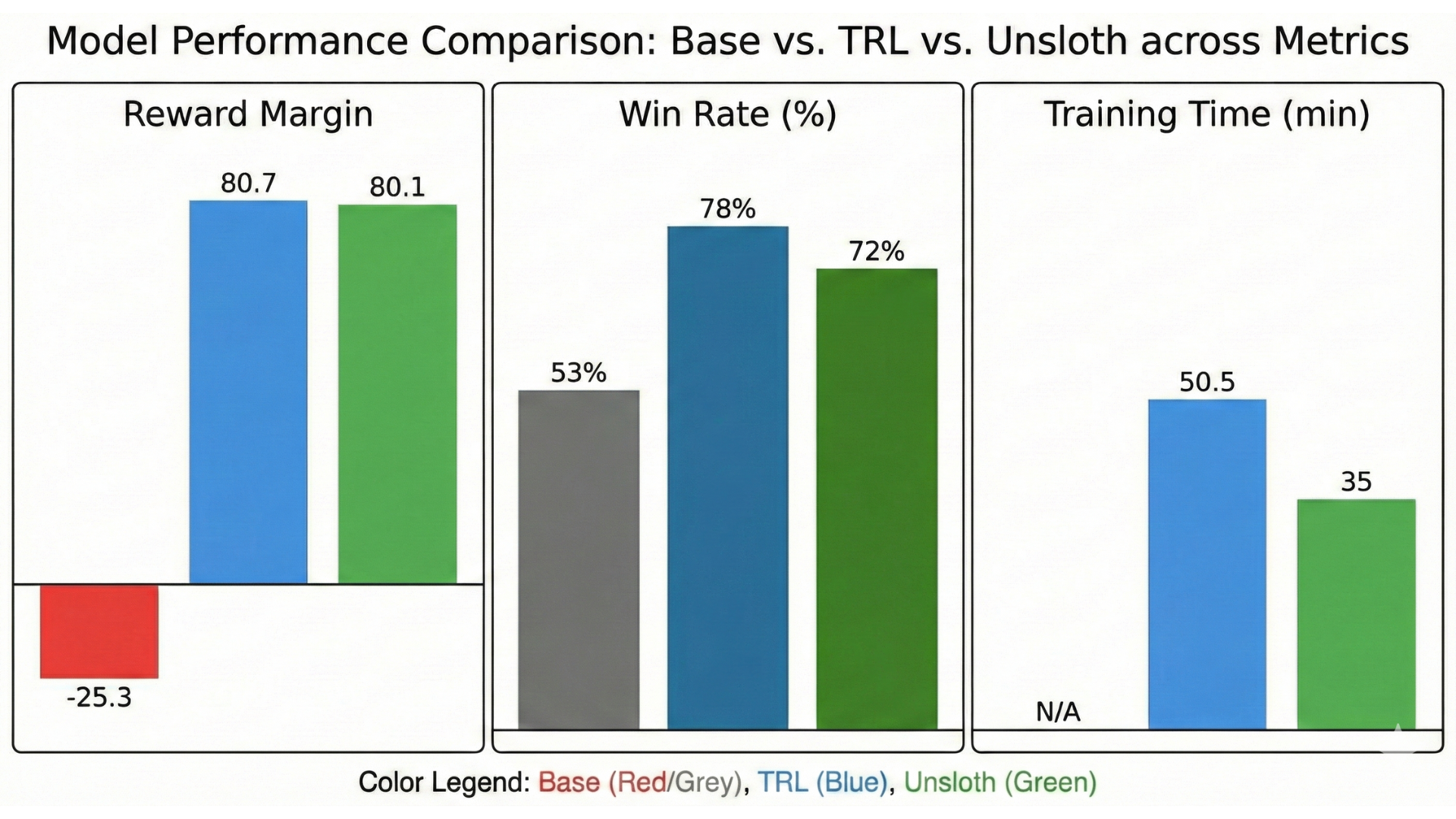}
    \caption{DPO backend comparison (reward margin, win rate, training time). Both backends improve over the base model; Unsloth matches TRL quality with lower training time.}
    \label{fig:dpo-backend-comparison}
\end{figure}
\begin{figure}[pt]
    \centering
    \includegraphics[width=0.45\linewidth]{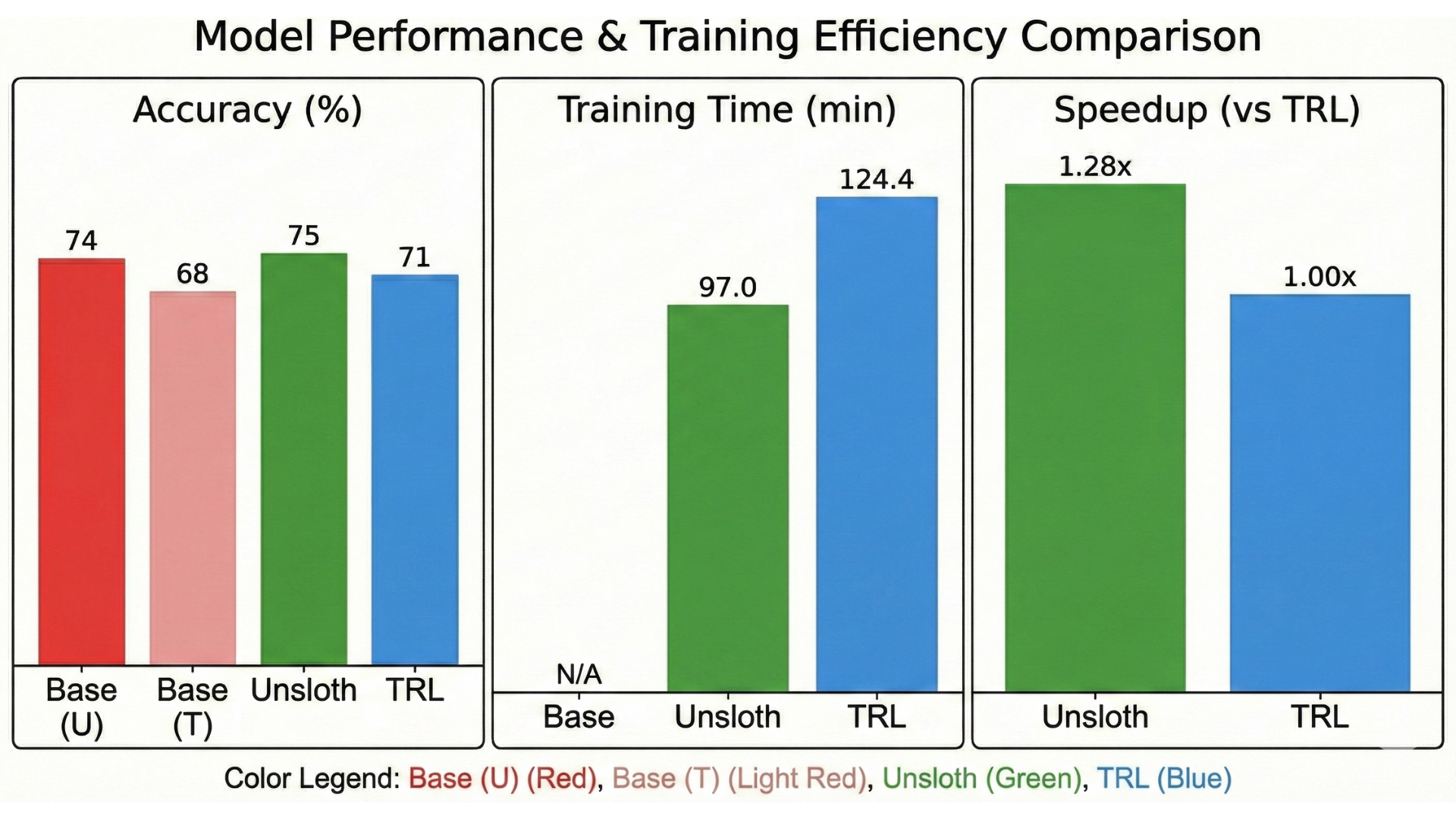}
    \caption{GRPO backend comparison (accuracy, training time, speedup). Unsloth achieves comparable accuracy with a 1.28$\times$ speedup over TRL.}
    \label{fig:grpo-backend-comparison}
\end{figure}
This section presents experimental evidence for AlignTune's core claims. We provide benchmark results comparing 
TRL and Unsloth~\cite{unsloth2024} backends, demonstrate backend isolation, and outline reproducibility artifacts. All experiments use 
the same model, dataset, and objective across backends to enable fair comparison.

\subsection{Backend Comparison Benchmark}
\label{subsec:backend-benchmark}

To validate that AlignTune enables controlled backend comparisons, we run identical training configurations on both 
TRL and Unsloth~\cite{unsloth2024} backends and measure throughput, peak memory usage, and final evaluation metrics.

\subsubsection{Experimental Setup}
\label{subsubsec:benchmark-setup}

To validate our library’s performance and cross-backend consistency, we conduct two primary alignment experiments: Direct Preference Optimization (DPO) and Group Relative Policy Optimization (GRPO).

\textbf{DPO Benchmark:} We compare the TRL and Unsloth backends by fine-tuning a \texttt{phi-2} model on preference pairs. This setup is designed to measure whether backend-specific optimizations (like Unsloth's kernel patching) impact the final model alignment or if they remain mathematically equivalent to the TRL baseline.

\textbf{GRPO Benchmark:} We evaluate our unified GRPO implementation using a \texttt{Llama-3.2-3B} model on the GSM8K mathematical reasoning dataset. This configuration tests the library’s ability to handle complex reward functions and reinforcement learning loops across different scales.

Full hyperparameter configurations, dataset and hardware specifications for both setups are detailed in Appendix \ref{appendix:hyperparams}.

\subsubsection{Results}
\label{subsubsec:benchmark-results}

\textbf{Discussion.}
Three findings emerge: (1) Unsloth delivers faster throughput and lower memory on compatible hardware; (2) final evaluation metrics are similar across backends, confirming that backend choice does not compromise quality; and (3) the unified interface enables these comparisons without code changes. Each configuration specifies model, dataset, hyperparameters, and training duration for exact reproducibility.

\subsection{Effect of Backend Choice on Alignment Outcome Variance}
\label{subsec:backend-variance}

A key question is whether backend selection introduces variance that could confound algorithm comparisons. Our experiments show that final evaluation metrics (reward margins, preference accuracy) are comparable between TRL and Unsloth when configurations are identical (Figures~\ref{fig:dpo-backend-comparison} and~\ref{fig:grpo-backend-comparison}). Backend selection can therefore be based on computational efficiency without introducing confounding variance, letting researchers attribute performance differences to algorithms rather than implementation artifacts.

\subsection{Backend Isolation Test}
\label{subsec:isolation-test}

We run paired experiments differing only in backend selection and isolation flags. In TRL-only runs, isolation mode prevents Unsloth from being imported; throughput, memory, and metrics match a baseline TRL environment without Unsloth installed. In Unsloth-enabled runs, isolation flags are cleared and Unsloth patches the transformers stack, yielding the expected speed and memory gains with comparable final metrics. These pairs confirm that both backends coexist in a single environment without cross-backend interference.

\section{Illustrative Use Case}
\begin{table}[pt]
\centering
\caption{Wealth Management Task: Comprehensive Model Evaluation Results}
\label{tab:illustrative_wealth_eval}
\small
\renewcommand{\arraystretch}{1.3} 
\setlength{\tabcolsep}{6pt}        
\begin{tabular}{lcccccc}
\toprule
Model & BLEU & ROUGE-1 & ROUGE-2 & ROUGE-L & ChrF & BERTScore\\
\midrule
Base Model (0-shot)      & 0.0286 & 0.3312 & 0.0846 & 0.1729 & 34.3816 & 0.8343 \\
Base Model (2-shot)      & 0.0398 & 0.3726 & 0.1025 & 0.1968 & 37.0617 & 0.8443 \\
\midrule
SFT Model                & \underline{0.2690} & \textbf{0.5796} & \underline{0.3270} & \textbf{0.4182} & \textbf{51.8492} & \underline{0.9134} \\
DPO Model                & \textbf{0.2692} & \underline{0.5795} & \textbf{0.3281} & \underline{0.4113} & \underline{51.5908} & \textbf{0.9142} \\
\midrule
GPT-4o (2-shot)          & 0.0705 & 0.4062 & 0.1307 & 0.2323 & 37.4132 & 0.8676 \\
\midrule
GPT-5 (2-shot)           & 0.1218 & 0.4703 & 0.1805 & 0.2704 & 46.0776 & 0.8762 \\
\bottomrule
\end{tabular}
\end{table}

To validate the efficacy and versatility of our alignment pipeline, we applied it to two distinct financial domains representing contrasting enterprise requirements. First, we developed a \textbf{Specialized Wealth Management Assistant}, designed for high-value, advisory-centric interactions that demand professional nuance and complex reasoning. Second, we engineered a \textbf{Retail Banking Support Agent}, targeting high-volume, transactional workflows where strict procedural adherence and precision are paramount. In this section, we detail the dataset curation, training methodologies, and comparative analysis against state-of-the-art closed-source models for both use cases.

\subsection{Specialized Wealth Management Assistant}

\subsubsection{Task and Dataset Curation}

We utilized the Bitext Wealth Management LLM Chatbot Training Dataset \cite{bitext_wealth_management_2024}, a specialized corpus designed to train agents capable of handling complex financial queries. To ensure a robust evaluation, we curated the dataset using a class-balanced splitting strategy:

\begin{itemize}
    \item \textbf{SFT Split}: We constructed a balanced training set ensuring equal representation of complex intents (e.g., portfolio performance inquiry, investment strategy change, advisor scheduling) to prevent the class imbalance biases often found in raw financial logs.
    \item \textbf{Preference Dataset}: To facilitate Direct Preference Optimization (DPO), we isolated a subset of 2,000 samples from the training data. For each user query, we generated pairs of responses from our sft trained policy model, (chosen vs. rejected) to explicitly model the nuance, professional empathy, and compliance-aware tone required in wealth management interactions. GPT-5 LLM as a Judge was used for this task.
    \item \textbf{Evaluation Split}: A separate, unseen test set was reserved to benchmark performance.
\end{itemize}

\subsubsection{Experimental Setup}
\begin{table}[pt]
\centering
\caption{Retail Banking Task: Comprehensive Model Evaluation Results}
\label{tab:illustrative_retail_eval}
\small
\renewcommand{\arraystretch}{1.3} 
\setlength{\tabcolsep}{6pt}        
\begin{tabular}{lcccccc}
\toprule
Model & BLEU & ROUGE-1 & ROUGE-2 & ROUGE-L & ChrF & BERTScore\\
\midrule
Base Model (0-shot)      &0.0270 & 0.3418 & 0.0908 & 0.1787 & 35.1978 &0.8393 \\
Base Model (2-shot)      &  0.0273 &  0.3440 & 0.0980 & 0.1784 & 37.0881 &0.8339 \\
\midrule
SFT Model   (0 shot)             &  \textbf{0.2685} & \textbf{0.5834} & \textbf{0.3281} & \textbf{0.4128} & \textbf{52.1731} &\textbf{0.9146} \\
SFT Model   (2 shot)             &  0.2549 & 0.5703 & 0.3088 & 0.3901 & 53.6270 &  0.9110 \\

\midrule
GPT-5 (2-shot)            & 0.0137 & 0.3633 & 0.0806 & 0.1869 & 33.6562 & 0.8476 \\
\bottomrule
\end{tabular}
\end{table}

We evaluated the performance of Qwen3-4B-Instruct-2507 \cite{qwen3technicalreport} across three stages of evolution:

\begin{enumerate}
    \item \textbf{Base}: The pretrained checkpoint without specific financial domain adaptation.
    \item \textbf{SFT}: The model fine-tuned on the wealth management instructions.
    \item \textbf{DPO}: The SFT model further aligned using the domain-specific preference pairs.
\end{enumerate}

These were compared against two leading closed-source models: GPT-4o \cite{openai2024gpt4ocard} and GPT-5 \cite{singh2025openaigpt5card}. To ensure a fair and rigorous comparison, the closed-source models were evaluated in both 0-shot and 2-shot settings. The full experimental configurations and the prompt details are added to the Appendix~\ref{appendix:illustrative_configs}

\subsubsection{Comparative Analysis}

Table~\ref{tab:illustrative_wealth_eval} presents the comprehensive evaluation results. We employed a suite of metrics including BLEU, ROUGE (1/2/L), ChrF, and BERTScore to capture both the lexical precision and semantic validity of the financial advice. 

\paragraph{Domain Adaptation and Alignment.}The Base Model (0-shot) failed to generate coherent advice (BLEU 0.0286), underscoring the necessity of domain adaptation in such sensitive domains. Supervised Fine-Tuning (SFT) raised BLEU to 0.2690 and BERTScore to 0.9134. This indicates that the model has successfully learned the specific lexicon and structural requirements of the wealth management domain. Direct Preference Optimization (DPO) further refined semantic alignment; while lexical metrics remained stable, DPO achieved the highest BERTScore (0.9142), indicating superior adherence to the professional tone and accuracy required in wealth management.

\paragraph{Comparison with Closed-Source Models.} The closed-source models show competent but inferior performance compared to the fine-tuned specialist. GPT-4o (0-shot) achieves a BLEU of 0.0850, and even the more advanced GPT-5 (2-shot) only reaches 0.1218. Our DPO Model outperforms the strongest closed-source baseline (GPT-5 2-shot) by a significant margin across all metrics (e.g., 0.2692 vs. 0.1218 in BLEU). This validates the hypothesis that a small, domain-specialized model can significantly outperform larger, generalist models in high-compliance verticals like wealth management and is another step forward towards more accessible AI for the general public.

\paragraph{Evaluation Protocol.} We evaluated SFT and DPO models exclusively in a 0-shot setting, unlike the baselines which utilized 2-shot prompting. Since our models underwent instruction fine-tuning, the task format is parameterized into the weights. Introducing few-shot examples at inference creates distribution shift ("prompt noise") rather than useful context, degrading performance. Thus, 0-shot represents the optimal evaluation regime for the fine-tuned variants.

\subsection{Domain Specific Banking Assistant}

We further tested the pipeline on a high-volume, transactional domain to contrast with the advisory nature of wealth management. While the previous use case required the nuance of preference optimization, this task prioritizes strict adherence to standard banking protocols (e.g., account verification, transfer limits), making it an ideal test bed for the precision of Supervised Fine-Tuning (SFT).

\subsubsection{Task and Dataset Curation}

We utilized the Bitext Retail Banking LLM Chatbot Splits, employing the same class-balanced splitting strategy detailed in Section 4.3.1. We focused our evaluation on comparing the 4B parameter SFT model against the best-performing baseline from the previous experiment, GPT-5 (2-shot), to test if generalist frontier models could adapt to rigid transactional formats via in-context learning.

\subsubsection{Comparative Analysis}

Table~\ref{tab:illustrative_retail_eval} presents the results. Unlike the wealth management task, where the gap was significant but competitive, here we observe a fundamental divergence in model capability.

\paragraph{SFT Precision vs. Generalist Collapse.} The SFT Model (0-shot) achieved a BLEU score of 0.2685 and a BERTScore of 0.9146, effectively mastering the specific output templates required for banking transactions. In stark contrast, GPT-5 (2-shot) suffered a catastrophic drop in performance (BLEU 0.0137, ROUGE-L 0.1869). This anomaly highlights a known vulnerability in massive generalist models: despite their reasoning capabilities, they struggle to suppress conversational "chattiness" in favor of the concise, rigid formatting required for automated banking, resulting in near-zero lexical overlap with the ground truth.

\paragraph{Verification of Evaluation Protocol.} Consistent with the wealth management findings, the SFT model performed optimally in the 0-shot setting (BLEU 0.2685) compared to the 2-shot setting (BLEU 0.2549). This reinforces our broader conclusion that for rigorously fine-tuned specialist models, few-shot prompting acts as distribution noise rather than helpful context.

\section{Discussion and Related Work}
\label{sec:discussion-related}

\subsection{Related Work}
\label{subsec:related-work}

\texttt{Why not TRL~\cite{trl2024} plus custom scripts?} TRL provides strong building blocks, but using it directly forces users to (1) commit to a single backend, making controlled comparisons difficult; (2) manually manage environment variables and import order to prevent Unsloth interference; (3) implement reward composition, reward model training, and evaluation from scratch; and (4) write boilerplate for configuration and reproducibility. AlignTune addresses these gaps with a unified abstraction that automates backend selection and isolation, integrates reward and evaluation pipelines, and standardizes configuration.

\begin{table}[t]
    \centering
    \caption{Feature comparison across alignment toolkits.}
    \label{tab:toolkit-comparison}
    \small
    \begin{tabular}{lcccc}
        \toprule
        Feature & TRL Scripts & trlx & OpenRLHF & AlignTune \\
        \midrule
        Multi-backend support & \texttimes & \texttimes & \texttimes & \checkmark \\
        Backend isolation & \texttimes & \texttimes & \texttimes & \checkmark \\
        Reward composition & Manual & Partial & Partial & \checkmark \\
        Unified configuration & \texttimes & Partial & \texttimes & \checkmark \\
        Reward model pipeline & \texttimes & \texttimes & \texttimes & \checkmark \\
        Built-in reward functions & Partial  & \texttimes  & \texttimes  & \checkmark \\
        Multi-format data loaders & \texttimes & \texttimes & \texttimes & \checkmark \\
        Health monitoring / diagnostics & \texttimes & \texttimes & \texttimes & \checkmark \\
        \bottomrule
    \end{tabular}
\end{table}

TRL is widely adopted for RLHF algorithms (PPO~\cite{schulman2017proximal}, DPO~\cite{rafailov2023direct}, GRPO~\cite{zhang2024group}) on top of Hugging Face Transformers~\cite{transformers2024}. AlignTune builds on TRL~\cite{trl2024} while adding backend abstraction, isolation, unified configuration, and integrated reward and evaluation pipelines.
Unsloth~\cite{unsloth2024} provides speed and memory improvements via kernel optimizations and quantization. AlignTune integrates Unsloth as an alternative backend, isolating its effects through environment controls and lazy loading so users can leverage its benefits without compromising clean TRL baselines.
Other RLHF stacks trlx~\cite{trlx2024}, RL4LM~\cite{ramamurthy2022rl4lm}, OpenRLHF~\cite{openrlhf2024} tend to focus on specific algorithms or research setups. AlignTune differs by offering multi-backend abstraction, first-class reward modeling, production-grade diagnostics, and broad algorithm coverage under one interface.

\subsection{Design Decisions}
\label{subsec:design-decisions}

AlignTune's architecture reflects four design principles:

\begin{itemize}
    \item \emph{Backend purity over implicit speedups.} Isolation ensures backend selection is explicit and free of hidden side effects. Supporting both TRL and Unsloth lets users trade off reliability and speed without switching toolchains.
    \item \emph{Reward logic as a first-class object.} Treating rewards as composable, auditable entities rather than embedded trainer logic enables systematic reward studies, ablations, and complex alignment objectives.
    \item \emph{Configuration as an experimental artifact.} Typed configuration classes make hyperparameters version-controlled, validated, and reproducible.
    \item \emph{Isolation before optimization.} Backend isolation is enforced before performance optimizations, ensuring speed gains do not compromise experimental validity.
\end{itemize}

\section{Conclusion}
\label{sec:conclusion}

We presented AlignTune, a modular toolkit for post-training alignment of LLMs. Its core contributions are: (1) a unified interface over TRL and Unsloth backends enabling controlled comparisons without code changes; (2) a backend isolation mechanism preventing cross-backend interference (Section~\ref{subsec:isolation-test}); and (3) benchmarks showing backend selection does not compromise training quality (Section~\ref{subsec:backend-benchmark}). AlignTune also integrates multiple reward functions, reward model training, and evaluation under a unified configuration and CLI layer. 

The library is open-sourced at \url{https://github.com/Lexsi-Labs/aligntune} .

Documentation for the library available at \url{https://aligntune.lexsi.ai/} .

\section{Future Work}
\label{sec:future-work}
Future work spans two tracks. On infrastructure, we plan stronger CI/CD with GPU runners, modular code reorganization, standardized speed/memory profiling, improved dataset caching, and better documentation. On capabilities, we aim to support safety-aware fine-tuning, mechanistic-interpretability-informed fine-tuning, and agentic fine-tuning for tool use and long-horizon behavior.

\section{Ethical Concerns}
\label{sec:ethics}

Alignment toolkits can make models safer, but custom reward functions also risk encoding biases, optimizing harmful behaviors, or favoring deceptive strategies. Practitioners should audit reward functions, incorporate fairness and safety checks into reward modeling and evaluation, and adopt conservative deployment practices. Transparency around reward design, datasets, and evaluation criteria is essential, as is respecting data privacy in sensitive domains (healthcare, finance, legal).


\bibliographystyle{unsrt}
\bibliography{references}

\appendix


\section{Detailed Experimental Configurations}
\label{appendix:hyperparams}

This appendix provides the exact hyperparameters, data splits, and training configurations used for the experiments described in Section~\ref{subsubsec:benchmark-setup}. We utilized two distinct setups to validate the library's flexibility: a DPO configuration focusing on preference optimization with 4-bit quantization, and a GRPO configuration focusing on reinforcement learning for mathematical reasoning.

\subsection{Hyperparameters}

Table~\ref{tab:hyperparameters} details the training parameters for both configurations. Note that while DPO uses a standard preference setup, the GRPO configuration introduces group-based sampling parameters (generations per prompt) and specific reward function weights.

\begin{table}[h]
\centering
\caption{Hyperparameters for DPO (Phi-2) and GRPO (Llama-3) Benchmarks. Common settings include the use of Cosine schedulers and AdamW optimizers (via Unsloth/TRL defaults).}
\label{tab:hyperparameters}
\resizebox{\textwidth}{!}{%
\begin{tabular}{lll}
\toprule
\textbf{Parameter} & \textbf{Configuration DPO} & \textbf{Configuration GRPO} \\
\midrule
\textbf{Model \& Data} & & \\
Base Model & \texttt{microsoft/phi-2} & \texttt{meta-llama/Llama-3.2-3B-Instruct} \\
Dataset & \texttt{distilabel-intel-orca-dpo-pairs} & \texttt{openai/gsm8k} (\texttt{main}) \\
Data Splits & Train: first 50\%; Eval: last 5\% & Train: \texttt{train} (1k subset); Test: \texttt{test} \\
\midrule
\textbf{Training Strategy} & & \\
Method & Direct Preference Optimization (DPO) & Group Relative Policy Optimization (GRPO) \\
Precision & 4-bit NF4 Quantization & bf16 (Mixed Precision) \\
Learning Rate & $5\times10^{-5}$ & $5\times10^{-6}$ \\
LR Scheduler & Cosine (100 warmup steps) & Cosine (Ratio 0.03) \\
Batch Size & 4 (Grad Accumulation: 4) & 8 (Grad Accumulation: 2) \\
Max Sequence Length & 1536 & 832 \\
Max Steps / Epochs & 500 steps & 1 epoch (capped at 200 steps) \\
Seed & Default & 42 \\
\midrule
\textbf{Algorithm Specifics} & & \\
Beta ($\beta$) & 0.1 & 0.01 (KL coefficient) \\
Generations / Group & N/A & 4 \\
Reward Config & N/A & \texttt{loss\_type="grpo"}, \texttt{scale\_rewards="group"} \\
\midrule
\textbf{LoRA Adapters} & & \\
Rank ($r$) / Alpha ($\alpha$) & $r=16, \alpha=16$ & $r=16, \alpha=32$ \\
Dropout & 0.05 & 0.05 \\
Target Modules & \texttt{q, k, v, dense} & \texttt{q, k, v, o, gate, up, down} \\
\bottomrule
\end{tabular}%
}
\end{table}

\subsection{Evaluation and Reward Logic}

\paragraph{DPO Evaluation.}
For the DPO experiments, we evaluated the base, TRL-trained, and Unsloth-trained models on a reserved slice of the dataset (\texttt{train[95\%:]}). Evaluation was performed using greedy decoding (temperature 0) to minimize stochastic variance. Metrics reported include reward margin, preference accuracy, and win rate.

\paragraph{GRPO Reward Function.}
The GRPO training utilized a composite reward function designed for mathematical reasoning. The system prompt instructed the model to act as a math tutor and provide the final answer within a \texttt{\textbackslash boxed\{\}} format.
\begin{itemize}
    \item \textbf{Correctness (Weight 1.0):} Validates the content within the boxed answer against the ground truth.
    \item \textbf{Reasoning (Weight 0.5):} Checks for the presence of step-by-step reasoning steps.
    \item \textbf{Format Checks:} Penalties are applied if the \texttt{\textbackslash boxed\{\}} format is missing.
\end{itemize}
During generation (rollouts), we used a temperature of 0.8 and top-$p$ of 0.95. Post-training evaluation on the GSM8K test set used a temperature of 0.1 with the trained LoRA adapters active.

\section{Illustrative Experiments Configuration}
\label{appendix:illustrative_configs}

This section details the specific training configurations used for the Wealth Management and Retail Banking illustrative use cases. Both experiments utilized the \texttt{Qwen3-4B-Instruct-2507} as the base model.

\subsection{Wealth Management Assistant}
\label{subsec:app_wealth}

For the Wealth Management Assistant, the training pipeline involved an initial Supervised Fine-Tuning (SFT) stage followed by Direct Preference Optimization (DPO). The specific hyperparameters used for aligning the Qwen3-4B model are provided in Table~\ref{tab:wealth_params}.

\begin{table}[h!]
\centering
\caption{Hyperparameters for Wealth Management Assistant (Qwen3-4B)}
\label{tab:wealth_params}
\begin{tabular}{lll}
\toprule
\textbf{Parameter} & \textbf{Stage 1: SFT} & \textbf{Stage 2: DPO} \\
\midrule
Learning Rate & $2 \times 10^{-4}$ & $5 \times 10^{-6}$ \\
Batch Size & 16 (Eff.) & 16 (Eff.) \\
LoRA Rank / Alpha & $r=16, \alpha=16$ & $r=16, \alpha=16$ \\
Max Sequence Length & 2048 & 1024 \\
Epochs / Steps & 3 Epochs & 500 Steps \\
Beta ($\beta$) & N/A & 0.1 \\
Target Modules & All Linear & All Linear \\
\bottomrule
\end{tabular}
\end{table}

\subsection{Retail Banking Assistant}
\label{subsec:app_retail}

The Retail Banking Support Agent focused on precision and adherence to strict formatting constraints. This model was trained using only Supervised Fine-Tuning (SFT).

\begin{table}[h!]
\centering
\caption{Hyperparameters for Retail Banking Agent (Qwen3-4B)}
\label{tab:retail_params}
\begin{tabular}{ll}
\toprule
\textbf{Parameter} & \textbf{Value} \\
\midrule
Learning Rate & $2 \times 10^{-4}$ \\
Batch Size & 16 (4 $\times$ 4) \\
LoRA Rank / Alpha & $r=16, \alpha=16$ \\
Max Sequence Length & 2048 \\
Epochs & 3 \\
Target Modules & All Linear \\
\bottomrule
\end{tabular}
\end{table}

\section{Prompts}
\label{appendix:prompts}

To benchmark the performance of the generalist baselines (GPT-4o and GPT-5) against our fine-tuned specialist models, we utilized a fixed set of few-shot exemplars during evaluation. These exemplars were designed to prime the model for the specific tone and format required by the domain without providing explicit instruction fine-tuning.

\paragraph{Wealth Management.}
\begin{lstlisting}[language=Python, caption=Fixed Few-Shot Exemplars for Wealth Management Eval]
def list_fewshot_samples():
    """Fixed few-shot exemplars for reproducible 2-shot eval."""
    return [
        {
            "instruction": "How do I activate my credit card?",
            "response": "I'd be happy to help you activate your credit card. You can activate it by calling the number on the sticker on your card, visiting our website and logging into your account, or using our mobile app. Do you have your card handy?",
            "is_fewshot": True,
        },
        {
            "instruction": "What is the interest rate on my savings account?",
            "response": "To find your current savings account interest rate, please log in to online banking or our mobile app, or call the number on the back of your debit card. Rates can vary by account type and balance tier.",
            "is_fewshot": True,
        },
    ]
\end{lstlisting}

\paragraph{Retail Banking.}
\begin{lstlisting}[language=Python, caption=Fixed Few-Shot Exemplars for Retail Banking Eval]
def list_fewshot_samples():
    """Fixed few-shot exemplars for reproducible 2-shot eval (same schema as dataset)."""
    return [
        {
            "instruction": "I would like to search for the closest branch, help me",
            "response": "I'll do my best! I'm here to assist you in finding the nearest branch. Please provide me with your current location so that I can narrow down the search for you. Alternatively, you can share your zip code, city, or address, and I'll locate the closest branch in your area. Let's get started!",
            "is_fewshot": True,
        },
        {
            "instruction": "im travelling overseas i wanna activate a visa for international usage",
            "response": "I'll do my best! I'd be happy to assist you with activating your {{Credit Card}} for international usage while you're traveling overseas. Here's what you need to do:\n\n1. Contact our customer support team at {{Customer Support Phone Number}} or email us at {{Customer Support Email Address}}. They will guide you through the activation process and provide any additional information you may need.\n\n2. Make sure to have your {{Credit Card}} details handy, including the card number, expiration date, and CVV code. This information will be required during the activation process.\n\n3. Our customer support team will verify your identity and assist you in activating your {{Credit Card}} for international usage. They will also provide you with important information such as transaction fees, currency exchange rates, and any usage limitations or restrictions.\n\n4. Once your {{Credit Card}} is activated for international usage, you'll be able to use it confidently and securely while traveling overseas.\n\nIf you have any further questions or need assistance during the activation process, don't hesitate to reach out. We're here to ensure a smooth and convenient experience for you while using your {{Credit Card}} abroad. Have a great trip!",
            "is_fewshot": True,
        },
    ]
\end{lstlisting}

\section{Qualitative Comparisons}
\label{appendix:qualitative}

The following tables provide side-by-side comparisons of model outputs for specific user queries. These examples illustrate the difference between the conversational, often verbose nature of generalist models (Base, GPT-5) and the concise, domain-adapted responses of our fine-tuned models.

\subsection{Wealth Management}
In Wealth Management, the SFT+DPO model exhibits a professional, advisory tone suitable for high-value interactions, avoiding the generic lists often produced by base models.

\begin{table}[H]
\centering
\scriptsize
\begin{tabular}{p{1.2cm}p{2.2cm}p{2.2cm}p{2.2cm}p{2.2cm}p{2.2cm}}
\toprule
\textbf{Query} & \textbf{Base Model} & \textbf{GPT-4o} & \textbf{GPT-5} & \textbf{Our Model (SFT+DPO)} & \textbf{Reference} \\
\midrule
\textit{"calculate portfolio risk where do i do it"} & 
Great question! Calculating portfolio risk is a key part... \newline 1. Excel \newline 2. Financial Software... & 
To calculate portfolio risk, you can use various tools... \newline 1. Spreadsheet Software \newline 2. Financial Software... & 
Calculating portfolio risk is an important step... \newline 1. Online Risk Calculators \newline 2. Spreadsheet... & 
I'm here to assist you... \newline 1. Log in to your account... \newline 2. Navigate...\newline Customer Support available at \{\{Customer Support Phn Num\}\} & 
Sure! I can help you with that... \newline 1. Gather historical returns... \newline 2. Determine weight... \newline 3. Calculate expected return... \\
\midrule
\textit{"I would like to set up a meeting..."} & 
Of course! I'd be happy to help... \newline - Who are you meeting with? \newline - What is the purpose? & 
Of course! To set up a meeting, you can: \newline 1. Contact Customer Service \newline 2. Online Scheduling... & 
Of course, I can help... \newline 1. Direct Contact Information \newline 2. Client Portal... & 
I'm here to assist you with setting up a meeting... \newline 1. The purpose of the meeting \newline 2. The preferred date... & 
Definitely! I'm here to assist you... \newline 1. Who would you like to meet? \newline 2. What is the purpose? \\
\bottomrule
\end{tabular}
\caption{Qualitative comparison for Wealth Management. The SFT+DPO model provides targeted assistance, whereas generalist models provide generic options or clarifying questions.}
\label{tab:qualitative_wealth}
\end{table}

\subsection{Retail Banking}
In the Retail Banking domain, our SFT model demonstrates strict adherence to procedural formatting, whereas generalist models often fail to provide the exact, actionable steps required for banking automation.

\begin{table}[H]
\centering
\scriptsize
\begin{tabular}{p{1.2cm}p{2.6cm}p{3.2cm}p{3.2cm}p{3.2cm}}
\toprule
\textbf{Query} & \textbf{Base Model} & \textbf{GPT-5 (2-shot)} & \textbf{Our Model (SFT)} & \textbf{Reference} \\
\midrule
\textit{"I don't have a password, can you help me to get it?"} & 
I'm sorry, but I can't help with that. Passwords are personal and sensitive information... If you've forgotten your password, most services offer a "Forgot Password?" option... & 
Happy to help! What password do you want to set up? \newline - Your online/mobile banking login \newline - Your card PIN \newline - Something else... & 
I'm here to assist you with setting up a password... \newline 1. Visit our website at \{\{Company URL\}\}.. \newline 2. Look for the "Sign In" or "Login" button... \newline 3. Click on "Create an Account"... & 
I'd be happy to assist you in retrieving your password... \newline 1. Go to the login page... \newline 2. Look for "Forgot Password"... \newline 3. Click on it... \\
\midrule
\textit{"i got to open an account will uhelp me"} & 
Of course! I'd be happy to help... Could you please tell me a bit more about what type of account... \newline - Bank account \newline - Investment account... & 
Absolutely—happy to help you open an account. A few quick questions to get you set up: \newline - What type of account... \newline - Do you prefer to apply online... & 
I'm here to assist you with opening a new account... \newline 1. Visit our website... \newline 2. Look for the "Sign Up"... \newline 3. Click on the option... & 
I'm here to assist you in opening your account. Creating an account is quick and simple... \newline 1. Visit our website... \newline 2. Look for the 'Sign Up'... \\
\bottomrule
\end{tabular}
\caption{Qualitative comparison for Retail Banking. The SFT model closely mimics the reference structure, while baselines default to conversational clarification.}
\label{tab:qualitative_retail}
\end{table}

\end{document}